\documentclass[10pt, conference, compsocconf]{IEEEtran}

\ifCLASSINFOpdf
\else
\fi

\usepackage{amssymb}

\usepackage{tablefootnote}

\usepackage{lscape}
\usepackage[pdftex]{graphicx}
\graphicspath{{./images/}}
\usepackage{subfig}
\usepackage[cmex10]{amsmath} 
\usepackage{footnote}
\makesavenoteenv{tabular}
\makesavenoteenv{table}

\usepackage{url} 
\usepackage{cite}
\usepackage{wasysym}

\begin{document}
\title{Redefining Binarization and the Visual Archetype}

\author{
\IEEEauthorblockN{Anguelos Nicolaou\IEEEauthorrefmark{1}
,Marcus Liwicki\IEEEauthorrefmark{2}}
\\
 \vspace{-7pt}
    \IEEEauthorblockA{
    \IEEEauthorrefmark{1}Computer Vision Center, Edifici O, Universitad Autonoma de Barcelona,Bellaterra, Spain\\
    \IEEEauthorrefmark{2} Document, Image and Voice Analysis (DIVA) Group, University of Fribourg - Switzerland\\
    Email: anguelos.nicolaou@cvc.uab.es, marcus.liwicki@unifr.ch}
    \\
}

\maketitle

\begin{abstract}
Although binarization is considered passe, it still remains a highly popular research topic.
In this paper we propose a rethinking of what binarization is.
We introduce the notion of the visual archetype as the ideal form of any one document.
Binarization can be defined as the restoration of the visual archetype for a class of images.
This definition broadens the scope of what binarization means but also suggests ground-truth should focus on the foreground.
\end{abstract}

\subsection{Defining Binarization}

Binarization has almost become a dirty word in some parts of the Document Image Analysis (DIA) community.
In the authors opinion the reason for this is not that it has been solved, not that it is an ill posed problem~\cite{lopresti2011problem} although it is, as much as the fact that the assumption of high quality binarization is practically infeasible in many real world cases.
Binarization is also usually required by methods that use smearing, morphological operations etc.
Furthermore in a recent experimental evaluation of the contribution methods have across a full DIA pipeline by Laminroy et al.~\cite{lamiroy2011document}, a binarization method proved to be the one with the highest positive contribution.

Some attempts at a definition of binarization have been made in the literature. 
Kavalieratou describes it as \textit{a method that discriminates foreground from background, thus, removing any kind of noise that obstructs the legibility of the document image}~\cite{kavallieratou2006improving}.
Stathis et al. defined it as automatically converting the document images in a bi-level form in such way that the foreground information is represented by black pixels and the background by white ones in ~\cite{stathis2008evaluation} and  Shafait et al. give a very similar definition~\cite{shafait2008efficient}.
Shafait et al. define binarization 
Ntirogiannis et al. define it as "the process that segments the document image into text and background by removing any existing degradations"~\cite{ntirogiannisCombinedBinarisation} .
What is more indicative, is 
It is indicative that several high impact papers addressing binarization as their main topic, describe it as \textit{selecting a threshold}~\cite{gatos2004adaptive}, or even wisely avoid defining at all~\cite{sauvola1997adaptive}.
In~\cite{lopresti2011problem} Lopresti and Nagy use binarization as an example of an ill-defined problem.
In recent years the DIBCO~\cite{gatos2009icdar} competitions have become the standard for benchmarking binarization methods.
By defining the problem with respect to a given ground-truth the discussion of what is binarization has been bypassed.
Smith et al. in~\cite{smith2012effect} address the question of bias in ground-truthing and propose monitoring the effect ground-truth has over method development.
What is not addressed with benchmarking on specific datasets, is the extent to which a method needs tuning.
\\

\begin{figure*}
\includegraphics[width=\textwidth]{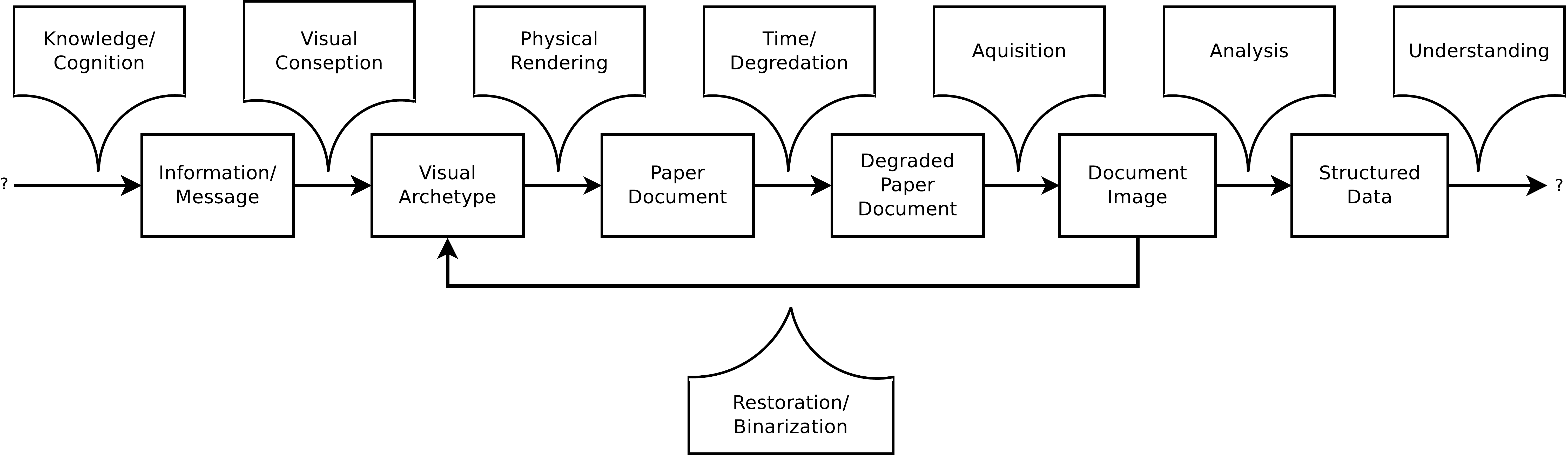}
\label{fig:definition_flowchart}
\caption{Proposed modelling of degradations and binarization}
\end{figure*}

\subsection{Visual Archetype}
In order to redefine binarization, we introduce the concept of the visual archetype.
The visual archetype is the ideal, intended, immaterial form of the document.
Depending on the modality, handwritten, typeface printing, desktop publishing, etc., the visual archetype is different things.
In all cases, the visual archetype, is the most mature, the last form the document takes in the production process before it goes on to the materialization stage.
In the desktop publishing case, the visual archetype would be the vector form of the document; the raster version of the document, if such is rendered, we consider part of the material stages of production in the sense that discretization is strongly influenced by the computational resources available etc.
In the typeface printing modality we consider the visual archetype to be the result of the typesetting process, what is codified on the compositing stick.
In the handwriting modality, the visual archetype is harder to identify.
The reason for the handwriting complication is that when one writes, he gets immediate feedback on how his text is rendered on paper.
We could hypothesize that the feedback is taken into account and if what was written diverged from the visual archetype, the visual archetype is adapted to incorporate what was written.
A scientific testing of such a hypothesis goes beyond the scope of this paper and might be unimportant, since our hypothesis was introduced to describe a concept and not a phenomenon.
Even though there is probably no specific visual archetype for handwriting in the scope of a full document or a page, the concept can make sense on a smaller scale such as a stroke, a letter, or a word.
We could maybe state that each grapheme is produced in the context of a specific visual archetype.
We suggest that declination from the visual archetype happens because of a lack of skill or external derogation such as ink drops, crumbled paper, etc..
Skill is required for visualizing a feasible archetype, for anticipating how graphemes will be rendered on paper, but also for executing the visual archetype with precision.
Taking the above assumptions, one could conclude that the more skilled a writer is, the more consistent during the writing of the document the visual archetype will be.
As an abstract concept, we associate the visual archetype with the document creator's intention and define it as \textbf{the most precise form a document creator's intentions take}.
The visual archetype's definition allows us to consider any deviation from it as a kind of imperfection, error, or, noise.
Note that visual archetypes should be put into reference with the work on cognitive models and prototype theory, i.e. they can be defined by idealized conceptual models, such as the typeface printing modality and the desktop printing modality, but are not limited to them.
Furthermore, the visual archetypes could be a kind of idealized visual model, i.e. the idea of the visual appearance in mind which should be put on paper. This can be put into relation with the idealized conceptual model (ICM) of Lakoff ~\cite{lakoff1999cognitive}, however a detailed analysis of this idea is beyond the scope of this paper.


\subsection{Two Tone Assumption}

If we address the question of how colors are modelled in visual archetypes an interesting dichotomy arises.
The visual archetypes for some document element types define colors in continuous ranges, while in other document element types as selections from a finite palette which in some cases contains only tones.
The authors of this paper postulate that the visual archetypes of text and drawings\footnote{A picture, image, etc., that is made by making lines on a surface with a pencil, pen, marker, chalk, etc., but usually not with paint. source: http://www.merriam-webster.com/dictionary/drawing , accessed: 2014-02-28} traditionally have discrete colors and in the overwhelming majority only two tones.
In support of our postulation we can only provide some relevant remarks.
In\cite{bringhurst1996elements} Bringhurst, in reference to the engraving the printing process produces on the surface of the paper, states that \textit{although early renaissance typographers were excited by the depth and the sense of touch they could achieve by the printing process, following neoclassical typographers such as Baskerville, would go as a far as to apply a process similar to ironing clothes on the printed document to produce perfectly flat surfaces}.
We suggest that text, even in the case that it wasn't, evolved to be bi-tonal because of the used media.
Liquid ink on paper, which dominated writing for centuries, makes the color tones practically uncontrollable to the layman.
When using a soft pen/plume, pressure variations result in line thickness variations.
This way handwriting can transmit non transcribable meanings in the spatial domain.
The bold and italic fonts could be described as descendants of the meanings that were associated with spatial information.
Contemporary font systems such as TTF contain a purely spatial description of the fonts; tone related modifications, such as anti-aliasing, are automatically inferred during rendering.
How could a gray-level document be transcribed without modern technology? What kind of document is untranscriptable?

\subsection{Document Image Binarization}

Using the above context, we call the inference of the visual archetype as document image restoration.
Binarization is therefore defined to be \textbf{the inference of the visual archetype for elements we assume had only two tones}.
In Fig.~\ref{fig:definition_flowchart} an abstract pipeline depicting briefly the creation of a document, followed by its analysis and how image restoration and binarization operate can be seen.
This definition of binarization does not restrict binarization methods to methods that produce binary maps of the image.
Since the visual archetype is closely related to the creation process of the document, the special case were the same tone is used as foreground in one region and as background in the other.
Any method that produces a continuous map of the image which is interpreted as inference of the two tone visual archetype falls also into the definition.
This extension of the definition introduces the notion of soft and hard binarization methods depending on the type of output they produce.
Based on the pipeline in Fig.~\ref{fig:definition_flowchart}, we could go as far as

\subsection{Complications}

Medieval manuscripts are a typical example of documents who's final form was produced in several iterations, in many cases each iteration by a different creator. We refer to such cases as incremental documents.
Medieval manuscripts contain scholia, corrections, and various forms of annotations which are put on paper independently of the original transcription.
In some cases, the layout of pages was defined and put in paper before and independently of the following transcription.
In a more modern example, students annotate lecture handouts~\cite{seuret2014}.
Such documents can be described as documents of an incremental nature.
and suggest a stack of several visual archetypes, one for each increment-stage of the document.
As in handwriting, we assume that during conception, each visual archetype incorporates the present state of the document.
The question how is the existing document represented in the visual archetype, if answerable, definitely goes beyond the present discourse.
It is safe to assume however that two visual archetypes of a document with an incremental nature can disagree, since the later archetype has incorporated imperfections in the rendering of the former.
In the special case where a document or a part of it were created by using ink for the background and its absence for a foreground, this should be modelled as an incremental document which has two layers, a background and a foreground.
Depending on the task at hand, we might aim at restoring different archetypes in the stack; in such a context, binarization becomes the inference or restoration of the last and cumulative visual archetype in that stack.

\subsection{Implications}

Adopting  the concept of visual archetypes under the two tone assumption presents repercussions on the topics of ground-truthing and performance evaluation in several tasks of DIA.
This implies that the ground-truth should be composed by classifying the components of the image if a binarization is feasible.
Ground-truthing of incremental documents can be aimed at the full stack of visual archetypes as those happened during the evolution of the document or a single archetype.
Yet, the fact that the ground-truth must not introduce visual information which is not in the visual archetype it approximates, remains.
From a data representation perspective, it is the authors opinion that encoding the groundtruth as labeled pixels in an discrete image or in a stack of discrete images is more consistent with the notion of the visual archetype.



\subsection{Conclusion}
In conclusion, in this paper we suggested a perspective that allows to rethink binarization and its role in DIA pipelines.
Whether implicit or explicit, binarization is part of most DIA pipelines.
If we assume document analysis to be the inverse of document synthesis, it only makes sense to start the analysis from where the synthesis left off and that is the visual archetype.
From the offered perspective only one imperative is arising: ground-truth of document image segmentations such as textline, word, character etc., should only annotate the foreground pixels.

\section*{Acknowledgement}

The authors would like to acknowledge Basilis G. Gatos, Dimosthenis Karatzas, Konstantinos Ntirogiannis, and Nicole Eichenberger for discussions and their input which was instrumental to the writing of this paper.


\bibliographystyle{IEEEtran}
\bibliography{IEEEabrv,qualitative_binarization_bibliography}

\end{document}